\documentclass{article}
\usepackage[margin=1in]{geometry}
\usepackage{graphicx} 
\usepackage[parfill]{parskip}
\usepackage[colorlinks, urlcolor=blue!80!black, linkcolor=blue!80!black, citecolor=green]{hyperref}
\usepackage{courier}
\usepackage{makecell}
\usepackage{xcolor}
\usepackage{array}
\usepackage{multirow}

\title{DANA: Domain-Aware Neurosymbolic Agents \\ for Consistency and Accuracy}

\author{
  Vinh Luong\textsuperscript{†}, Sang Dinh\textsuperscript{†}, Shruti Raghavan\textsuperscript{†}, William Nguyen\textsuperscript{†}, Zooey Nguyen\textsuperscript{†}, \\ Quynh Le\textsuperscript{†}, Hung Vo\textsuperscript{†}, Kentaro Maegaito\textsuperscript{§}, Loc Nguyen\textsuperscript{†}, Thao Nguyen\textsuperscript{†}, Anh Hai Ha\textsuperscript{†}, \\ Christopher Nguyen\textsuperscript{†} \\
  \texttt{\{vinh,sang,shruti,william,zooey,}\\
  \texttt{quynh,hung,loc,thao,annieha,ctn\}@aitomatic.com}\\
  \texttt{maegaito@fenrir.co.jp}
  \\
  \\
  \textsuperscript{†}Aitomatic, Inc. \\
  \textsuperscript{§}Fenrir, Inc.
}

\date{}

\begin{document}

\maketitle

\begin{abstract}
Large Language Models (LLMs) have shown remarkable capabilities, but their inherent probabilistic nature often leads to inconsistency and inaccuracy in complex problem-solving tasks. This paper introduces DANA (Domain-Aware Neurosymbolic Agent), an architecture that addresses these issues by integrating domain-specific knowledge with neurosymbolic approaches. We begin by analyzing current AI architectures, including AutoGPT, LangChain ReAct and OpenAI's ChatGPT, through a neurosymbolic lens, highlighting how their reliance on probabilistic inference contributes to inconsistent outputs. In response, DANA captures and applies domain expertise in both natural-language and symbolic forms, enabling more deterministic and reliable problem-solving behaviors. We implement a variant of DANA using Hierarchical Task Plans (HTPs) in the open-source OpenSSA framework. This implementation achieves over 90\% accuracy on the FinanceBench financial-analysis benchmark, significantly outperforming current LLM-based systems in both consistency and accuracy. Application of DANA in physical industries such as semiconductor shows that its flexible architecture for incorporating knowledge is effective in mitigating the probabilistic limitations of LLMs and has potential in tackling complex, real-world problems that require reliability and precision.
\end{abstract}


\section{Background \& Motivation}

With the rise of large language models (LLMs) as a potent new means for organizing, retrieving and reasoning about diverse information, various attempts have been made to use them for autonomous problem-solving. AutoGPT\cite{autogpt_github} is an early implementation in this category and has become one of the most followed open-source AI projects since its inception in 2023. Many popular LLM frameworks and ecosystems such as LangChain\cite{LangChain_github} have followed suit, creating their own agent implementations such as LangChain ReAct\cite{LangChain_react}.

So far, the “agent bloom” has borne some productive fruits, especially in creative content-generation applications. However, there has not been meaningful success in real-world industry domains that require reliability and precision in solving complex problems, especially when dealing with physical processes and operations. Practitioners have encountered inconsistency and inaccuracy when applying autonomous agents to use cases in which they need determinism and reproducibility and not creativity. These issues stem from LLMs’ inherent probabilistic nature, which allows for flexibility and generalization but introduces unpredictability in outputs.

This paper investigates the fundamental causes of the encountered inconsistency and inaccuracy through examining current AI-agent system architectures’ uses of neural and symbolic components. Based on such analysis, we propose DANA (Domain-Aware Neurosymbolic Agent), an architecture that incorporates domain-specific knowledge and more symbolic structure to enhance deterministic behaviors and improve consistency and accuracy in complex problem-solving tasks.

The remainder of this paper is organized as follows: Section 2 analyzes current AI system architectures through a neurosymbolic lens, identifying their strengths and limitations. Section 3 introduces the DANA (Domain-Aware Neurosymbolic Agent) architecture that addresses the identified limitations. Section 4 details the implementation of DANA in the OpenSSA framework and presents empirical benchmarking results on the FinanceBench dataset\cite{Islam2023FinanceBenchAN}, demonstrating DANA’s superior performance in both consistency and accuracy. Section 5 describes a use case employing DANA to advise on semiconductor manufacturing process recipes, highlighting its practical applicability in industrial workflows. Section 6 discusses DANA’s implications for industrial AI system designs, particularly the role of symbolic structures and operations, the role of domain-specific knowledge and the potential for automated knowledge-to-agent systems. Finally, Section 7 concludes the paper, summarizing our contributions and outlining future research directions.

\section{AI System Architectures Through a Neurosymbolic Lens}

To solve a meaningfully complex problem, an AI system needs to perform two principal activities: \textbf{program search}, in which the system comes up with a program for solving the posed problem, and \textbf{program execution}, in which the system runs that program for the solution. The involved program may take various forms, from natural language to symbolic representations such as graphs, hierarchies, or programming-language code.

\subsection{Review of Recent Planning \& Reasoning Literature}

Recent literature on LLM-based planning and reasoning has created many techniques implemented in autonomous agent frameworks. An early approach, and one that continues to inspire new adaptations, is Chain of Thought (CoT)\cite{Wei2022Chain}. Zero-shot CoT, also known as single-path reasoning, is effective for straightforward problems but is limited in its capacity to synthesize multiple reasoning paths for the best result. Advancements such as Chain of Thought with Self-Consistency (CoT-SC)\cite{Wang2022Self-Consistency} address this limitation by reasoning through multiple paths and employing majority voting to determine the most consistent output. The Tree of Thought (ToT)\cite{Yao2023Tree} framework expands on CoT-SC by considering planning as a search problem. This approach represents each potential step in a plan as a node in a tree structure. At each step, models evaluate all possible subsequent steps and choose the most appropriate one.

The LLM-only CoT/ToT approaches have been hindered by LLMs’ inconsistency, and hybrid approaches have been proposed, incorporating traditional symbolic planners into the planning and reasoning process. Techniques such as LLM+P \cite{Liu2023LLM+P} and LLM-DP \cite{Dagan2023Dynamic} involve translating natural language into symbolic representations, which are then processed by external planners to aid in decision-making. Although these methods show promise in enhancing planning capabilities, they remain contingent on LLMs’ proficiency in accurately converting natural language into symbolic forms. Furthermore, the generated plans may not always align with human preferences.

Researchers have also increasingly recognized the importance of additional feedback and/or input from the environment, auxiliary models and human experts or users. For instance, ReAct \cite{Yao2022ReAct} relies on environmental feedback after an action is taken, using it to reason and generate subsequent steps. LLM-Planner \cite{Song2022LLM-Planner} employs physical grounding to create and update plans for embodied agents. Voyager \cite{Wang2023Voyager} enhances its planning by incorporating feedback from program execution progress, errors and self-verification results, which refine future actions. Similarly, Ghost \cite{Zhu2023Ghost} integrates environmental feedback, such as action outcomes and state changes, into its reasoning and decision-making processes. SayPlan \cite{Rana2023SayPlan} uses a scene graph simulator to derive feedback, iteratively adjusting its strategies based on outcomes and state transitions until a successful plan is formed. SelfCheck \cite{Miao2023SelfCheck} enables models to evaluate their own step-by-step reasoning, identifying and correcting errors. InterAct  \cite{Chen2023InterAct} uses different models to assist the main model in avoiding errors and unnecessary actions. Still, both environmental and model-based feedback have limitations, particularly in tasks requiring output consistency, due to the inherent probabilistic behaviors of the models themselves. Human feedback, as leveraged by Inner Monologue \cite{Huang2022Inner} through human interactions, shows potential for producing model outputs more consistent and well-aligned with human preferences.

Overall, when greater control over the system’s output is desired, researchers are shifting from LLM-only to neurosymbolic conceptual frameworks and technical implementations, with a growing recognition for the importance of human input.

\subsection{Review of Recent Neurosymbolic Literature}

Researchers have created techniques that combine the strengths of neural networks and symbolic reasoning to achieve better-controlled outcomes. Xu et al. ~\cite{Xu2017ASL} used symbolic structures as a form of regularization during the training phase of neural networks. This regularization acts as a bridge between neural output vectors and predefined logical constraints, ensuring that the outputs align more closely with logical expectations. Such approaches have been particularly beneficial in domains requiring visual reasoning~\cite{Mao2019Neuro-Symbolic}, where they enable models to represent complex structures like scenes, words and sentences with a symbolic concept learner. This combination facilitates more accurate visual understanding and reasoning by integrating symbolic reasoning capabilities with the perceptual power of neural networks. In addition to visual reasoning, neurosymbolic AI has been applied to algorithmic reasoning tasks, particularly in scenarios involving symbolically grounded inputs~\cite{Ebrahimi2021TowardsBT, Ebrahimi2021TowardsBT, Velivckovic2022TheCA}. Another approach involves generating outputs that are consistent with a predefined symbolic knowledge base~\cite{Manhaeve2018DeepProbLogNP, Marra2019IntegratingLA, Zhang2023TractableCF}, thereby addressing the uncertainty and variability inherent in purely neural approaches. By grounding neural network outputs in a symbolic framework, these approaches reduce the ambiguity of the results and ensure adherence to logical rules.

LLMs have further enabled the combination of the flexibility of neural networks with the rigorous structure of symbolic reasoning, thanks to their ability to generate symbolic representations in formal languages such as Planning Domain Definition Language (PDDL) \cite{Liu2023LLM+P} and First-Order Logic (FOL) \cite{Olausson2023LINCAN}. These representations can serve as inputs into traditional symbolic planners \cite{Olausson2023LINCAN} or differentiable symbolic programming frameworks~\cite{Zhang2023ImprovedLR}. The typical pipeline of translating natural language into logical languages, performing symbolic reasoning and then translating the results back into natural language allows for a more interpretable and controllable AI system, in which the reasoning process is transparent and can be audited or refined. Logic-LM~\cite{Pan2023LogicLMEL} is one of the most notable implementations of this approach. However, a key challenge is formulating correct and effective logical representations for complex problems. Inaccurate formulations lead to inefficiencies and errors in reasoning, as symbolic reasoners struggle to resolve ambiguities or inconsistencies. To address this, Logic-LM++~\cite{Kirtania2024LOGICLMMR} introduces a process for the system to iteratively refine logical formulations. However, despite these neurosymbolic advancements, without human guidance AI systems still face many challenges in complex, domain-specific problems that require deep expert knowledge.

\subsection{Analyses of Current AI Architectures}

\subsubsection{AutoGPT and Related Frameworks \& Tools}

AutoGPT is an open-source autonomous agent implementation. Given a problem to solve, an AutoGPT agent creates and then executes a list of tasks for doing so. Whenever it cannot accomplish a task right away, it further creates and executes a decomposed list of sub-tasks. At the individual task level, in addition to processing textually described tasks, an AutoGPT agent can also trigger commands – interchangeably called “actions” or “tools” – which involves getting or creating and then running Python code for various purposes, including querying information from external sources (e.g., Google Search), manipulating files and exporting data back to the user.

AutoGPT has inspired a number of derivative or similar autonomous agent implementations, including AgentGPT~\cite{agentgpt_github} and GodmodeAI~\cite{godmode_space}, which add functionalities such as data-source connections, human-in-the-loop approval and feedback, and handling of non-text data such as images.

Interpreted through our neurosymbolic program-search-and-program-execution paradigm, the AutoGPT family creates problem-solving programs by an LLM, without reference to domain-specific knowledge. During program execution, textually described tasks are processed by the LLM, tool-calling tasks are processed by their tool runners (external APIs or pre-written Python functions), and Python-code tasks are run by a Python interpreter.

Developers have used AutoGPT quite well for tasks whose results can be open-ended, such as internet research, creative content creation, and business analysis. The developer community around AutoGPT has also made progress in identifying good prompt engineering practices to get helpful outputs~\cite{autogpt_github}.

Despite AutoGPT’s potential, developers have run into significant limitations. The fully decomposed, nested lists of tasks are typically quite large, hence AutoGPT is usually resource- and time-intensive. AutoGPT also often gets stuck in loops, and its performance may also be unpredictable due to varying task complexity, making its real-world use limited to the range of tasks that are slightly too tedious for humans but not so complex that it would get stuck somewhere. AutoGPT’s performance in high-precision tasks such as coding has been underwhelming~\cite{autogpt_github}, leading to ongoing discussions about enhancing its efficacy~\cite{autogpt_github}.

\subsubsection{LangChain ReAct}

LangChain ReAct, another open-source autonomous agent implementation, combines the reasoning-and-acting (ReAct) conceptual framework \cite{Yao2022ReAct} and the LangChain ecosystem of tools. A LangChain ReAct agent solves a problem by iterative CoT reasoning with evidence collected from the agent’s available tools.

Like AutoGPT, LangChain ReAct creates problem-solving programs by an LLM without reference to domain-specific knowledge. During program execution, textually described tasks are processed by the LLM, while tool-calling tasks are processed by the corresponding LangChain tools – which are all Python functions, some of which may trigger neural (LLM) processing such as retrieval-augmented generation (RAG). There is currently no affordance for on-the-fly programming-code creation and execution.

LangChain ReAct performs better than typical question-answering engines and RAG agents when multiple information sources and multiple reasoning passes are needed to arrive at a solution. However, its users have also encountered issues of inconsistency~\cite{autogpt_github} and inaccuracy.

\subsubsection{OpenAI ChatGPT-4o \& Assistant API}

OpenAI’s ChatGPT-4o and its corresponding Assistant API exhibit materially improved problem-solving capability, especially in data analysis.

Although the technical architectures of OpenAI’s proprietary products are undisclosed, we could speculatively infer certain underlying components and working mechanisms through observing how ChatGPT-4o works. When faced with a complex data-analysis problem, ChatGPT-4o seemingly creates a solution plan that includes Python coding tasks and integrative reasoning tasks. The Python code is created by the GPT-4o LLM and run by a Python interpreter. The plans are likely quite compact – each analysis typically involves up to five code snippets, and seemingly only a couple of integrative reasoning tasks.

ChatGPT-4o has a basic Memory integration that can help retain important facts stated by users. Users can tactically use this Memory mechanism to store relevant domain-specific knowledge, with the hope of enhancing ChatGPT-4o’s response quality. Whether and how Memory is leveraged by ChatGPT-4o’s in data analysis is unclear.

Reports of ChatGPT-4o’s inconsistency and and inaccuracy suggest its architecture is also not rigorously equipped to handle complex analytical scenarios.

\subsection{Summary and Takeaways from Current Architectures}

The above analyses of current AI architectures are summarized in Table \ref{current-ai-architectures}.

\renewcommand{\arraystretch}{1.5} 

\begin{table}[ht]
\centering
\begin{tabular}{|>{\raggedright\arraybackslash}p{2.5cm}|>{\raggedright\arraybackslash}p{3cm}|>{\raggedright\arraybackslash}p{5cm}|>{\raggedright\arraybackslash}p{4cm}|}
\hline
\textbf{Architecture} & \textbf{Program Search} & \textbf{Task Formats \& Execution} & \textbf{Knowledge Use} \\ \hline
AutoGPT family & Program created by LLM (\textit{neural}), usually quite large & 
\textbf{Textually-described tasks}: processed by LLM (\textit{neural}) \newline\newline
\textbf{Tool-calling tasks}: run by external APIs or Python functions (\textit{symbolic}), with argument values provided by LLM (\textit{neural}) \newline\newline
\textbf{Python code}: run by Python interpreter (\textit{symbolic}) & none \\ \hline

LangChain ReAct & Program created by LLM (\textit{neural}) & 
\textbf{Textually-described tasks}: processed by LLM (\textit{neural}) \newline\newline
\textbf{Tool-calling task}: run by Python functions (\textit{symbolic}, which may in turn trigger \textit{neural}), with argument values provided by LLM (\textit{neural}) & none \\ \hline

OpenAI ChatGPT-4o \& Assistant API & Program created by LLM (\textit{neural}), seemingly quite shallow/small & 
\textbf{Textually described tasks}: processed by LLM (\textit{neural}) \newline\newline
\textbf{Python code}: run by Python interpreter (\textit{symbolic}) & Knowledge can be injected into Memory, but how Memory is used is unclear \\ \hline
\end{tabular}
\caption{Current AI system architectures through a neurosymbolic lens}
\label{current-ai-architectures}
\end{table}

This reveals several common patterns explaining inconsistency and inaccuracy in complex problem-solving:
\begin{enumerate}
    \item Reliance on LLMs for program creation from scratch, leading to variability in problem solving approaches;
    \item Heavy involvement of LLMs in program execution, contributing to inconsistency at the individual task level; and
    \item Limited or no incorporation of domain-specific knowledge relevant to problem-solving.
\end{enumerate}

These weaknesses suggest that AI-system architectures could be improved by shifting more responsibility to symbolic structures and their more deterministic operations, and by organizing and utilizing relevant domain-specific knowledge.

\section{DANA: Domain-Aware Neurosymbolic Agent}

Based on the above insights, we propose DANA (Domain-Aware Neurosymbolic Agent), an architecture paradigm designed to achieve greater deterministic behavior and ultimately more consistent and accurate output in autonomous problem-solving.

The DANA architecture is based on three core principles:

\begin{enumerate}
    \item First-class treatment of domain-specific knowledge;
    \item Use of both natural-language and symbolic representations of knowledge; and
    \item Explicit modeling of knowledge-capture (CAPTURE) and knowledge-application (APPLY) processes.
\end{enumerate}

These principles are realized through the following key components of DANA:

\begin{enumerate}
    \item \textbf{Knowledge Capture Process}: manual and automated methods for receiving, extracting and translation knowledge, and persisting such knowledge in the Knowledge Store and the Program Store for use in program search and program execution during problem-solving;
    \item \textbf{Knowledge Store}: a repository of domain-specific knowledge, including definitions, facts, formulas, expert heuristics and inference rules;
    \item \textbf{Program Store}: a storage of pre-existing programs applicable to well-characterized problems in the domain; and
    \item \textbf{Program Search Process with Finder \& Creator}: mechanisms for finding suitable pre-existing programs and creating new programs when no such are found.
\end{enumerate}

\begin{figure}[ht]
    \centering
    \includegraphics[width=\linewidth]{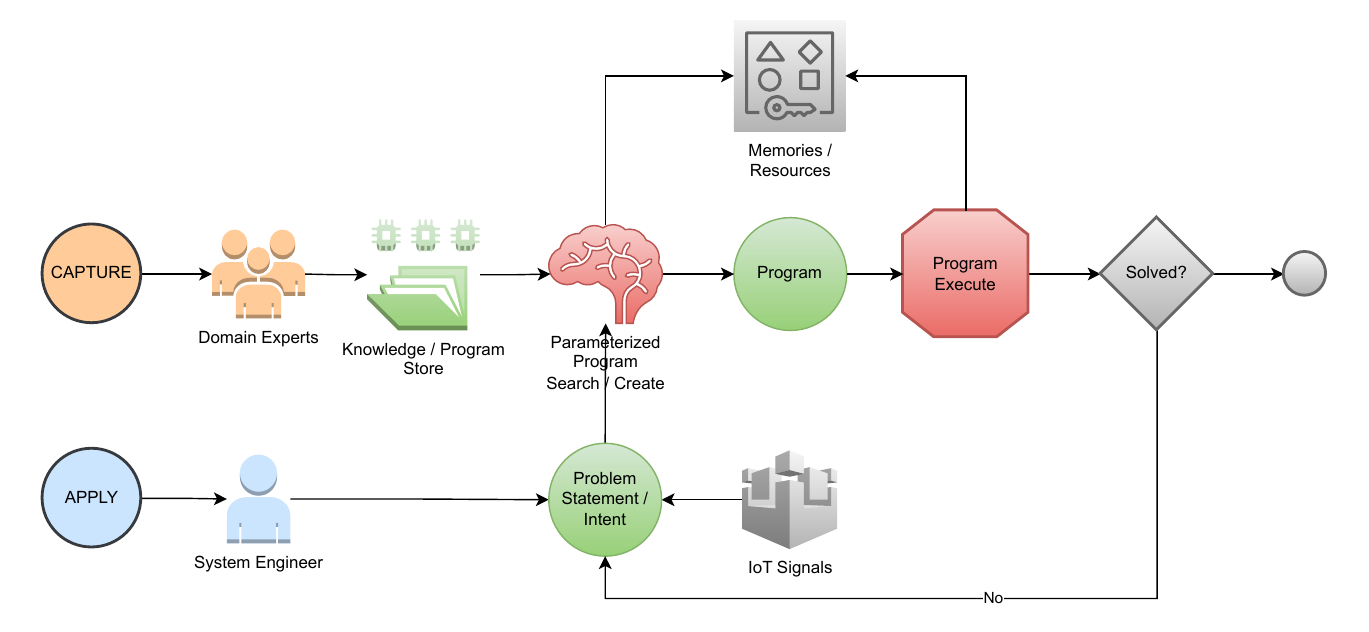}
    \caption{DANA Architecture. CAPTURE domain knowledge from experts. APPLY that to solve problems. Mediated by neurosymbolic building blocks.}
    \label{fig:DANA-components}
\end{figure}

These components work together per Figure~\ref{fig:DANA-components} to form a knowledge-first architecture that leverages domain expertise while maintaining the flexibility and power of neural approaches. DANA can be implemented in several variants, accommodating different knowledge capture mechanisms and program forms, allowing it to adapt to various problem domains and organizational needs.

In the following subsections, we elaborate on these components, their interactions, and how they collectively address the issues of inconsistency and inaccuracy.

\subsection{Knowledge Store}

The Knowledge Store is a collection of definitions, facts, formulas, expert heuristics and inference rules useful in the concerned domain. This knowledge can be referred to and leveraged in both program search and program execution, providing a foundation of domain-specific information.

\subsection{Program Store}

The Program Store is a storage of programs known to be applicable to certain well-characterized problems in the concerned domain. Each program comes with descriptive metadata about its purpose, facilitating the program search step during problem-solving. This component allows DANA to quickly identify and apply known solutions to familiar problem types.

\subsection{Program Search Process with Program Finder \& Program Creator}

The Program Finder is a mechanism for finding pre-existing programs applicable to posed problems, by looking into the Program Store and leveraging the domain knowledge in the Knowledge Store.

The Program Creator is a mechanism for creating new programs when no applicable pre-existing ones are found in the Program Store. It can leverage the domain knowledge in the Knowledge Store to create more effective and domain-appropriate programs.

\subsection{Knowledge Capture Process \& Knowledge Representations}

Because domain-specific knowledge significantly influences problem-solving quality, DANA treats the capture and representation of facts-and-rules knowledge and known programs as first-class concerns.

\subsubsection{Knowledge Capture}

DANA supports two main methods for populating the Knowledge Store and Program Store:

\begin{enumerate}
    \item \textbf{Manual Capture}: domain experts or AI engineers directly input knowledge and programs; and
    \item \textbf{Automated Capture} (AC): a Knowledge Encoding Assistant interviews domain experts and automatically encodes their knowledge and problem-solving strategies.
\end{enumerate}

\subsubsection{Facts-and-Rules Knowledge Forms}
The facts-and-rules knowledge in DANA can take two forms:

\begin{enumerate}
    \item \textbf{Natural-Language Knowledge} (NK): knowledge represented in natural language, easily understandable by humans; and
    \item \textbf{Symbolic Knowledge} (SK): knowledge represented in more formal, symbolic forms (e.g., Prolog relations).
\end{enumerate}

During problem-solving, NK is processed by neural engines (e.g., LLMs) for interpretation and reasoning, while SK is processed by symbolic engines with deterministic operations.

\subsubsection{Program Forms}
Likewise, known programs in DANA can also take two forms:

\begin{enumerate}
    \item \textbf{Natural-Language Programs} (NP): programs described largely in natural language, e.g., Hierarchical Task Plans (HTPs) with natural-language tasks; and
    \item \textbf{Symbolic Programs} (SP): programs in symbolic forms, such as Python code.
\end{enumerate}

\begin{table}[ht]
\centering

\begin{tabular}{|>{\raggedright\arraybackslash}p{4cm}|>{\centering\arraybackslash}p{4cm}|>{\centering\arraybackslash}p{4cm}|>{\centering\arraybackslash}p{4cm}|}
\hline

\multirow{2}{4cm}{\textbf{Knowledge Capture Mechanism}} & \multirow{2}{4cm}{\textbf{Facts-and-Rules Knowledge Form}} & \multicolumn{2}{c|}{\textbf{Program Form}} \\ \cline{3-4} 
 &  & Natural Language (NP) & Symbolic (SP) \\ \hline

\multirow{2}{4cm}{Manual} & Natural Language (NK) & DANA--NK--NP & DANA--NK--SP \\ \cline{2-4}
 & Symbolic (SK) & DANA--SK--NP & DANA--SK--SP \\ \hline

\multirow{2}{4cm}{Automated (AC)} & Natural Language (NK) & DANA--AC--NK--NP & DANA--AC--NK--SP \\ \cline{2-4}
 & Symbolic (SK) & DANA--AC--SK--NP & DANA--AC--SK--SP \\ \hline

\end{tabular}

\caption{DANA architecture variants}
\label{tab:dana-design-space}
\end{table}

\subsection{DANA Design Space}

By supporting the above alternative capture methods and knowledge and program forms, DANA can flexibly adapt to different problem domains. The above design choices and their various combinations make up the DANA design space described in Table \ref{tab:dana-design-space}. These permutations and hybrids combining them allow for flexible and versatile AI system designs incorporating diverse knowledge content and formats.

\section{Implementation \& Empirical Benchmarking}

To validate the efficacy of DANA, we implemented a specific variant in the open-source OpenSSA framework (\url{https://github.com/aitomatic/openssa/blob/main/openssa/core/agent/dana.py}) and compared its performance against those of current AI systems on a well-known problem-solving benchmark dataset.

\subsection{OpenSSA Implementation of DANA}

We implemented the DANA-NK-NP variant, which uses natural-language facts-and-rules knowledge representation and natural-language programs. Implementation details include:

\begin{enumerate}
\item \textbf{Knowledge Store}: a text storage containing domain-specific definitions, facts, formulas, expert heuristics, and rules of thumb, all described in natural language;
\item \textbf{Program Format}: Hierarchical Task Plans (HTPs) with natural-language task descriptions, which provide a human-understandable means for expressing how a goal can be achieved or a problem solved through hierarchical decomposition;
\item \textbf{Program Store}: a structured storage containing HTPs, each with a unique name and descriptive metadata about the problem it is designed to solve;
\item \textbf{Program Finder}: implemented with an LLM for recognizing directly applicable programs for posed problems through descriptive metadata in the Program Store;
\item \textbf{Program Creator}: implemented with an LLM that can be instructed to decompose a target problem or task into a more detailed HTP with sub-tasks; and
\item \textbf{Program Execution Mechanism}: utilizes Observe-Orient-Decide-Act reasoning (OODAR) for executing HTPs.
\end{enumerate}

\subsection{Empirical Benchmarking on FinanceBench}

\begin{table}[ht]
\centering
\begin{tabular}{|c|c|c|}
\hline
\textbf{Difficulty Level} & \textbf{\#Qs} & \textbf{Description} \\ \hline
0-RETRIEVE & 56 & Retrieve a single data point \\ \hline
1-COMPARE & 23 & Compare a small number of retrievable data points \\ \hline
2-CALC-CHANGE & 9 & Calculate relative change in same retrievable data point over time \\ \hline
3-CALC-COMPLEX & 43 & Calculate complex financial metrics involving multiple data points \\ \hline
4-CALC-AND-JUDGE & 10 & Calculate complex financial metrics and judge their goodness/healthiness \\ \hline
5-EXPLAIN-FACTORS & 2 & Explain major driving factors behind a change \\ \hline
6-OTHER-ADVANCED & 7 & Answer an unusually tricky financial question \\ \hline
\end{tabular}
\caption{FinanceBench difficulty levels}
\label{tab:difficulty_levels}
\end{table}

We evaluated our DANA implementation using the FinanceBench dataset, a collection of 150 financial-analysis cases based on public quarterly and annual financial reports filed with the US Securities and Exchange Commission (SEC), which we classified into increasing difficulty levels detailed in Table \ref{tab:difficulty_levels}.

\subsubsection{Experiment Setup}

We compared the performance of our DANA implementation against LlamaIndex RAG agents, LangChain ReAct agents and OpenAI Assistants, with each system having access to the same SEC filings and tasked to solve the same FinanceBench problems based on such filings (\url{https://github.com/aitomatic/openssa/tree/main/examples/FinanceBench}).

\subsubsection{Evaluation Metrics}

We evaluated the systems on two metrics:
\begin{enumerate}
    \item \textbf{Average Accuracy}: For each problem, we generated ten solutions from each AI system. We then calculated the percentage of solutions being correct per the FinanceBench ground truths. Performing such scoring across 150 FinanceBench cases and taking an average gave us an Average Accuracy score for the AI system.

    \item \textbf{Average Consistency}: From the same ten solutions generated for each problem from each AI system, we assigned a consistency score for that solution set based on whether those solutions were consistently correct or incorrect per the FinanceBench ground truths, with a 100\%-consistency score for an all-correct or all-incorrect solution set, and a 0\%-consistency score for a half-correct-and-half-incorrect solution set. Precisely, the consistency score is calculated as twice the absolute difference between 50\% and the proportion of the ten solutions being correct. Performing such scoring across 150 FinanceBench cases and taking an average gave us an Average Consistency score for the AI system.

\end{enumerate}

\subsubsection{Results}

\begin{figure}[ht]
    \centering
    \includegraphics[width=\linewidth]{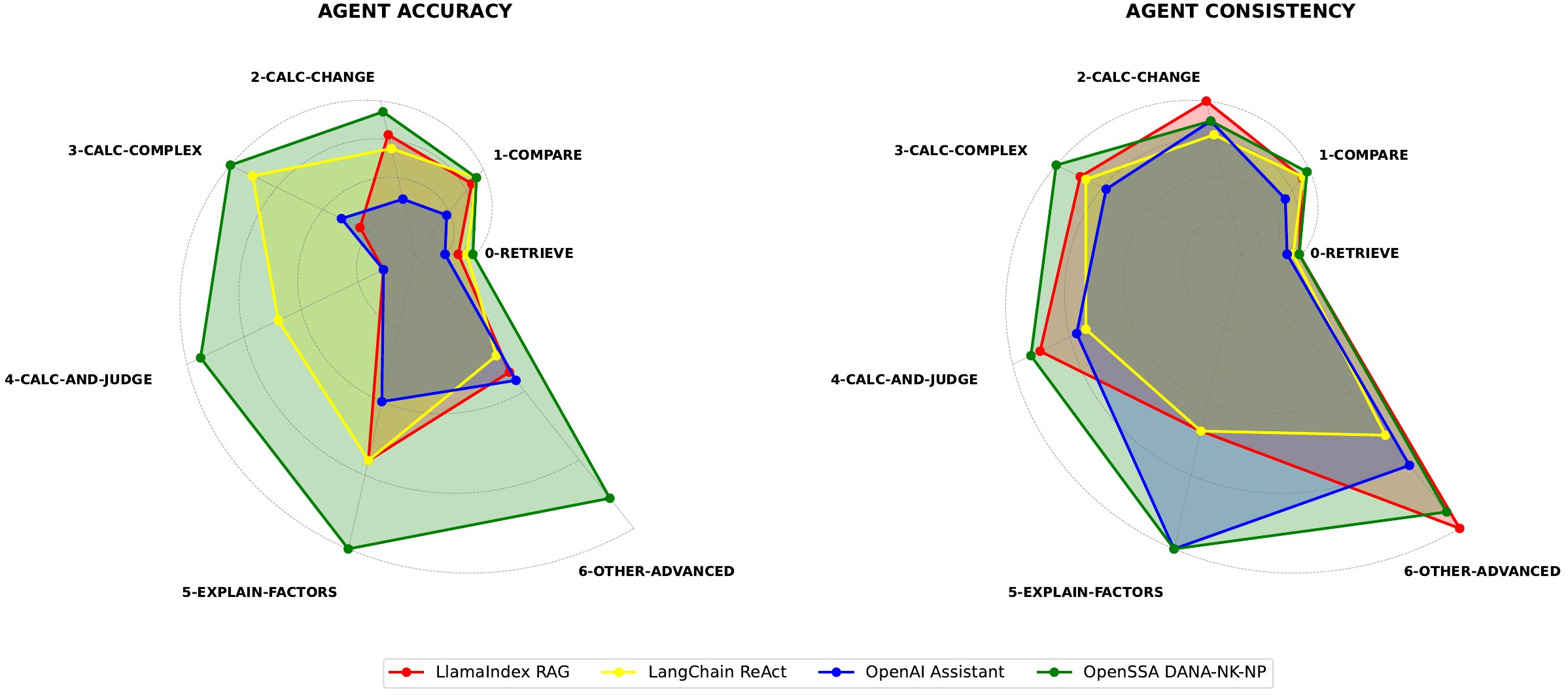}
    \caption{\textbf{Comparison of DANA and other agents.} OpenSSA DANA outperforms alternative agent frameworks
and tools on both Accuracy and Consistency.}
    \label{fig:DANA-comparison}
\end{figure}

\renewcommand{\arraystretch}{1.5} 

\begin{table}[ht]
\centering
\begin{tabular}{|>{\centering\arraybackslash}p{4cm}|>{\centering\arraybackslash}p{1.5cm}|>{\centering\arraybackslash}p{2cm}|>{\centering\arraybackslash}p{2cm}|>{\centering\arraybackslash}p{2cm}|>{\centering\arraybackslash}p{2cm}|}
\hline
\textbf{Difficulty Level} & \textbf{\#Qs} & \textbf{LlamaIndex RAG} & \textbf{LangChain ReAct} & \textbf{OpenAI Assistant} & \textbf{OpenSSA DANA-NK-NP} \\ \hline
0-RETRIEVE & 56 & 71\% & 85\% & 49\% & \textbf{95\%} \\ \hline
1-COMPARE & 23 & 83\% & \textbf{90\%} & 46\% & \textbf{90\%} \\ \hline
2-CALC-CHANGE & 9 & 78\% & 69\% & 36\% & \textbf{93\%} \\ \hline
3-CALC-COMPLEX & 43 & 31\% & 88\% & 40\% & \textbf{100\%} \\ \hline
4-CALC-AND-JUDGE & 10 & 14\% & 60\% & 14\% & \textbf{94\%} \\ \hline
5-EXPLAIN-FACTORS & 2 & 70\% & 70\% & 50\% & \textbf{100\%} \\ \hline
6-OTHER-ADVANCED & 7 & 43\% & 37\% & 46\% & \textbf{89\%} \\ \hline
\end{tabular}
\caption{\textbf{Average Accuracy} across FinanceBench difficulty levels (from 10 runs)}
\label{tab:accuracy}
\end{table}

\begin{table}[ht]
\centering
\begin{tabular}{|>{\centering\arraybackslash}p{4cm}|>{\centering\arraybackslash}p{1.5cm}|>{\centering\arraybackslash}p{2cm}|>{\centering\arraybackslash}p{2cm}|>{\centering\arraybackslash}p{2cm}|>{\centering\arraybackslash}p{2cm}|}
\hline
\textbf{Difficulty Level} & \textbf{\#Qs} & \textbf{LlamaIndex RAG} & \textbf{LangChain ReAct} & \textbf{OpenAI Assistant} & \textbf{OpenSSA DANA-NK-NP} \\ \hline
0-RETRIEVE & 56 & \textbf{96\%} & 86\% & 76\% & \textbf{96\%} \\ \hline
1-COMPARE & 23 & 90\% & 91\% & 65\% & \textbf{97\%} \\ \hline
2-CALC-CHANGE & 9 & \textbf{100\%} & 78\% & 87\% & 87\% \\ \hline
3-CALC-COMPLEX & 43 & 87\% & 84\% & 73\% & \textbf{100\%} \\ \hline
4-CALC-AND-JUDGE & 10 & 88\% & 68\% & 72\% & \textbf{92\%} \\ \hline
5-EXPLAIN-FACTORS & 2 & 60\% & 60\% & \textbf{100\%} & \textbf{100\%} \\ \hline
6-OTHER-ADVANCED & 7 & \textbf{100\%} & 66\% & 77\% & 94\% \\ \hline
\end{tabular}
\caption{\textbf{Average Consistency} across FinanceBench difficulty levels (from 10 runs)}
\label{tab:consistency}
\end{table}

The metrics in Tables~\ref{tab:accuracy} and~\ref{tab:consistency} and Figure~\ref{fig:DANA-comparison} demonstrate that, as a multi-step problem-solving tool, DANA is significantly more consistent than LangChain ReAct and OpenAI Assistant. The closest alternative in terms of consistency is LlamaIndex RAG, thanks to it being a one-pass information-retrieval-and-synthesization tool. DANA also outperforms in accuracy, helped by its strength in maintaining consistency through multiple steps in the problem-solving process, especially for complex problems (3-CALC-COMPLEX and above) which cannot be addressed with one-pass/one-step or few-step tools.

\section{DANA in Industrial Workflows: Semiconductor Case Study}

This section describes at a high level an application of DANA in a physical-industry workflow which requires that AI provide highly precise analyses and recommendations.

\subsection{Use Case Motivation: Semiconductor Etching Recipe Formulation}

The semiconductor industry has many complex design and manufacturing processes – e.g., design rule checking, layout generation, photolithography, etching, deposition, ion implantation, gate oxide formation, insulation, bonding, etc. – which need careful calibration to achieve optimal technical and economic outcomes. Etching recipe formulation is one such intricate task: it involves dozens of machine parameters that affect key physical operations and chemical reactions. A recipe lacking in rigor would result in inefficient cycle times or expensive reliability and quality problems such as etch rate anomalies, mask erosion, pattern distortion, plasma instability and non-uniformity, which negatively impact yield. Hence, the formulation of feasible and optimal etching recipes requires deep domain expertise and time-consuming expert analyses. There is consequently a great need for scaling up and making more available the knowledge of scarce experts in this domain.

\subsection{Expertise CAPTURE into DANA Knowledge and Program Stores}

Domain knowledge in etching was acquired from two principal sources: (i) discussions and interviews with etching experts at equipment makers and chip manufacturers, and (ii) public discussion forums in the semiconductor and related scientific research and engineering fields, such as ResearchGate. The knowledge was then organized into two storages that DANA agents could leverage:

\begin{enumerate}
    \item \textbf{Facts-and-Rules Knowledge Store:} facts about common gas types and their key properties, rules of thumb about correlations between etch rate changes and key process parameters, trade-offs across ranges of power and pressure parameter settings, and important safety advice to adhere to in etching operations; and
    \item \textbf{Program Store:} step-by-step and hierarchical expert procedures for analyzing the specifications, deciding on possible key parameter values, and judging the likely advantages and disadvantages of different parameter combinations in terms of etching outcomes such as plasma stability and uniformity.
\end{enumerate}

\subsection{Expertise APPLY through DANA-based Etching Advisor}

An Etching Advisor, an AI agent constructed with the DANA architecture, was given access to the above etching expert Knowledge and Program Stores. Additionally, this Etching Advisor employed SemiKong (\url{https://SemiKong.ai}), a semiconductor industry-specific LLM, in its program search and program execution, to further enhance the precision and relevance of its analyses and recommendations. This DANA-based Etching Advisor was then capable of providing operationally sound etching recipes, including pros-and-cons comparisons among feasible alternatives, thus helping process engineers save much analysis time and arrive at their final recipes more quickly. An example recommendation is presented in Figure~\ref{fig:DANA-demo}.

\begin{figure}[ht]
    \centering
    \includegraphics[width=\linewidth]{figs/DANA-demo.pdf}
    \caption{Typical recipe analysis \& recommendation from DANA-based Semiconductor Etching Advisor}
    \label{fig:DANA-demo}
\end{figure}

\section{Implications for Industrial AI System Designs}

This section explores DANA’s implications for the design of AI systems in industrial settings, focusing on the role of symbolic structures and operations, the role of domain-specific knowledge and the potential for automated knowledge-to-agent systems.

\subsection{Role of Symbolic Structures and Deterministic Operations}

One key contributor to DANA’s superior consistency is this architecture’s more explicit and greater assigned responsibility to symbolic elements in organizing domain-specific knowledge, and in structuring and executing problem-solving programs such as HTPs. The use of symbolic structures and their more deterministic operations significantly reduces the variability in individual processing step results and in how such results are aggregated or consolidated during an entire program’s execution.

As AI agents are increasingly employed in physical-industry applications, the stringent requirements on consistency shall likely further necessitate symbolic representations and deterministic processing. Current AI system architectures would need to evolve from their current predominant reliance on neural processing in order to be effective in such use cases.

\subsection{Role of Domain-Specific Knowledge}

First-class treatment of domain-specific knowledge, both in the facts-and-rules form and the program form, is another key driver of DANA’s superior consistency and accuracy in complex problem-solving. Such knowledge integration is crucial for success in industrial AI:

\begin{enumerate}
    \item \textbf{Reduced Critical Errors}: applications in high-stakes fields such as manufacturing and healthcare greatly need such consistency and accuracy to mitigate the risk of costly or dangerous accidents that can occur with purely data-driven approaches;
    \item \textbf{Improved Interpretability}: domain-specific knowledge representation makes AI decision-making processes more explainable, transparent and interpretable; and
    \item \textbf{Faster Deployment}: leveraging existing domain expertise can accelerate the deployment of AI systems in new industrial contexts, reducing the need for extensive data collection and model training to achieve comparable problem-solving quality.
\end{enumerate}

\subsection{Automated Knowledge-to-Agent Systems}

The promising performance of DANA points to the potential for developing automated systems that can transform domain knowledge into operational AI agents:

\begin{enumerate}
    \item \textbf{Scalable Expertise}: automated knowledge capture mechanisms could allow industries to scale their AI capabilities more efficiently, capturing and operationalizing the knowledge of multiple experts;
    \item \textbf{Continuous Learning}: automated systems  with active monitoring and evaluation could potentially update their knowledge bases in real-time, allowing industrial AI to adapt quickly to new information or changing conditions; and
    \item \textbf{Customization and Specialization}: the various DANA variants can be used for tailoring AI systems to specific industrial needs.
\end{enumerate}

\subsection{Challenges and Future Directions}

While DANA demonstrates significant potential, several challenges remain to be addressed:

\begin{enumerate}
    \item \textbf{Knowledge Elicitation}: developing effective methods for extracting and formalizing expert knowledge can be difficult, particularly in highly specialized industrial domains;
    \item \textbf{Handling Uncertainty}: further research is needed to enhance DANA’s capabilities in dealing with uncertain or incomplete information; and
    \item \textbf{Handling Contradictory Knowledge}: we need mechanisms to detect and handle contradictions in DANA’s Knowledge Store and Program Store – especially those containing large numbers of knowledge items or programs, among which contradictions may be indirect and nuanced – in order to safeguard problem-solving quality, especially in settings where such contradiction is costly or dangerous.
\end{enumerate}

\section{Conclusion}

This paper introduced DANA (Domain-Aware Neurosymbolic Agent), an architecture designed to address the inconsistency and inaccuracy in current LLM-based AI systems.  The key contributions of this work include:

\begin{enumerate}
    \item An analysis of current AI system architectures through a neurosymbolic lens, identifying limitations of purely probabilistic approaches
    \item DANA, a flexible architecture that combines domain expertise with neurosymbolic integration to enhance consistency and accuracy in problem-solving;
    \item An implementation of DANA in the open-source OpenSSA framework;
    \item Empirical evidence of DANA’s superior performance on the FinanceBench dataset over current AI systems;
    \item A case study of applying DANA in a high-stakes physical-industry process, namely semiconductor etching recipe formulation; and
    \item An analysis of DANA’s implications for industrial AI system designs, highlighting the importance of symbolic structures and operations, the value of domain-specific knowledge and the potential for automated knowledge-to-agent systems.
\end{enumerate}

Looking forward, DANA opens up several avenues for future research:

\begin{enumerate}
    \item Automated knowledge capture mechanisms to streamline the process of incorporating domain expertise into AI systems;
    \item Hybrid approaches that can dynamically balance between neural and symbolic processing based on the nature of the problem at hand;
    \item Extension of DANA to multi-domain settings, exploring how knowledge can be effectively transferred and applied across different areas of expertise; and
    \item More sophisticated neurosymbolic reasoning techniques that can handle uncertainty and incomplete information more effectively.
\end{enumerate}

As AI continues to proliferate in critical real-world industries, approaches like DANA that prioritize consistency, accuracy and domain-specific knowledge can play an important role in building effective and trustworthy AI systems.

\section*{Acknowledgments}

We would like to express our gratitude to the AI Alliance (\url{https://thealliance.ai}) for providing the impetus, resources, and platform for this work, and for collaboration in open science. In particular, we thank to the following members for their valuable contributions to this study:

\begin{itemize}
    \item \textbf{Adam Pingel}, IBM Head of Open Tools and Applications, The AI Alliance
    \item \textbf{Dean Wampler}, IBM Chief Technical Representative to The AI Alliance
    \item \textbf{Rick Johnson}, University of Notre Dame
    \item \textbf{Quan Dang}, FPT AI Center
    \item \textbf{Tinh Pham}, FPT AI Center
\end{itemize}

Their expertise, insights, and collaborative spirit have been instrumental in advancing our research.

\bibliographystyle{plain}
\bibliography{references}

\begin{thebibliography}{10}

\bibitem{agentgpt_github}
{\em AgentGPT}, 2024.

\bibitem{godmode_space}
{\em GodMode}, 2024.

\bibitem{Chen2023InterAct}
Po-Lin Chen and Cheng-Shang Chang.
\newblock Interact: Exploring the potentials of chatgpt as a cooperative agent.
\newblock {\em ArXiv}, abs/2308.01552, 2023.

\bibitem{Dagan2023Dynamic}
Gautier Dagan, Frank Keller, and Alex Lascarides.
\newblock Dynamic planning with a llm.
\newblock {\em ArXiv}, abs/2308.06391, 2023.

\bibitem{Ebrahimi2021TowardsBT}
Monireh Ebrahimi, Aaron Eberhart, Federico Bianchi, and Pascal Hitzler.
\newblock Towards bridging the neuro-symbolic gap: deep deductive reasoners.
\newblock {\em Applied Intelligence}, 51:6326 -- 6348, 2021.

\bibitem{autogpt_github}
Significant Gravitas.
\newblock Langchain react.
\newblock 2024.

\bibitem{Huang2022Inner}
Wenlong Huang, F.~Xia, Ted Xiao, Harris Chan, Jacky Liang, Peter~R. Florence, Andy Zeng, Jonathan Tompson, Igor Mordatch, Yevgen Chebotar, Pierre Sermanet, Noah Brown, Tomas Jackson, Linda Luu, Sergey Levine, Karol Hausman, and Brian Ichter.
\newblock Inner monologue: Embodied reasoning through planning with language models.
\newblock In {\em Conference on Robot Learning}, 2022.

\bibitem{Islam2023FinanceBenchAN}
Pranab Islam, Anand Kannappan, Douwe Kiela, Rebecca Qian, Nino Scherrer, and Bertie Vidgen.
\newblock Financebench: A new benchmark for financial question answering.
\newblock {\em ArXiv}, abs/2311.11944, 2023.

\bibitem{Kirtania2024LOGICLMMR}
Shashank Kirtania, Priyanshu Gupta, and Arjun Radhakirshna.
\newblock Logic-lm++: Multi-step refinement for symbolic formulations.
\newblock {\em ArXiv}, abs/2407.02514, 2024.

\bibitem{LangChain_github}
LangChain.
\newblock Langchain.
\newblock 2024.

\bibitem{LangChain_react}
LangChain.
\newblock Langchain react.
\newblock 2024.

\bibitem{Liu2023LLM+P}
B.~Liu, Yuqian Jiang, Xiaohan Zhang, Qian Liu, Shiqi Zhang, Joydeep Biswas, and Peter Stone.
\newblock Llm+p: Empowering large language models with optimal planning proficiency.
\newblock {\em ArXiv}, abs/2304.11477, 2023.

\bibitem{Manhaeve2018DeepProbLogNP}
Robin Manhaeve, Sebastijan Dumancic, Angelika Kimmig, Thomas Demeester, and Luc~De Raedt.
\newblock Deepproblog: Neural probabilistic logic programming.
\newblock {\em ArXiv}, abs/1907.08194, 2018.

\bibitem{Mao2019Neuro-Symbolic}
Jiayuan Mao, Chuang Gan, Pushmeet Kohli, Joshua~B. Tenenbaum, and Jiajun Wu.
\newblock The neuro-symbolic concept learner: Interpreting scenes, words, and sentences from natural supervision.
\newblock {\em ArXiv}, abs/1904.12584, 2019.

\bibitem{Marra2019IntegratingLA}
Giuseppe Marra, Francesco Giannini, Michelangelo Diligenti, and Marco Gori.
\newblock Integrating learning and reasoning with deep logic models.
\newblock {\em ArXiv}, abs/1901.04195, 2019.

\bibitem{Miao2023SelfCheck}
Ning Miao, Yee~Whye Teh, and Tom Rainforth.
\newblock Selfcheck: Using llms to zero-shot check their own step-by-step reasoning.
\newblock {\em ArXiv}, abs/2308.00436, 2023.

\bibitem{Olausson2023LINCAN}
Theo~X. Olausson, Alex Gu, Benjamin Lipkin, Cedegao Zhang, Armando Solar-Lezama, Josh Tenenbaum, and Roger Levy.
\newblock Linc: A neurosymbolic approach for logical reasoning by combining language models with first-order logic provers.
\newblock In {\em Conference on Empirical Methods in Natural Language Processing}, 2023.

\bibitem{Pan2023LogicLMEL}
Liangming Pan, Alon Albalak, Xinyi Wang, and William~Yang Wang.
\newblock Logic-lm: Empowering large language models with symbolic solvers for faithful logical reasoning.
\newblock {\em ArXiv}, abs/2305.12295, 2023.

\bibitem{Rana2023SayPlan}
Krishan Rana, Jesse Haviland, Sourav Garg, Jad Abou-Chakra, Ian~D. Reid, and Niko S{\"u}nderhauf.
\newblock Sayplan: Grounding large language models using 3d scene graphs for scalable task planning.
\newblock In {\em Conference on Robot Learning}, 2023.

\bibitem{Song2022LLM-Planner}
Chan~Hee Song, Jiaman Wu, Clay Washington, Brian~M. Sadler, Wei-Lun Chao, and Yu~Su.
\newblock Llm-planner: Few-shot grounded planning for embodied agents with large language models.
\newblock {\em 2023 IEEE/CVF International Conference on Computer Vision (ICCV)}, pages 2986--2997, 2022.

\bibitem{Velivckovic2022TheCA}
Petar Velivckovi'c, Adri{\`a}~Puigdom{\`e}nech Badia, David Budden, Razvan Pascanu, Andrea Banino, Mikhail Dashevskiy, Raia Hadsell, and Charles Blundell.
\newblock The clrs algorithmic reasoning benchmark.
\newblock In {\em International Conference on Machine Learning}, 2022.

\bibitem{Wang2023Voyager}
Guanzhi Wang, Yuqi Xie, Yunfan Jiang, Ajay Mandlekar, Chaowei Xiao, Yuke Zhu, Linxi~(Jim) Fan, and Anima Anandkumar.
\newblock Voyager: An open-ended embodied agent with large language models.
\newblock {\em Trans. Mach. Learn. Res.}, 2024, 2023.

\bibitem{Wang2022Self-Consistency}
Xuezhi Wang, Jason Wei, Dale Schuurmans, Quoc Le, Ed~Huai hsin Chi, and Denny Zhou.
\newblock Self-consistency improves chain of thought reasoning in language models.
\newblock {\em ArXiv}, abs/2203.11171, 2022.

\bibitem{Wei2022Chain}
Jason Wei, Xuezhi Wang, Dale Schuurmans, Maarten Bosma, Ed~Huai hsin Chi, F.~Xia, Quoc Le, and Denny Zhou.
\newblock Chain of thought prompting elicits reasoning in large language models.
\newblock {\em ArXiv}, abs/2201.11903, 2022.

\bibitem{Xu2017ASL}
Jingyi Xu, Zilu Zhang, Tal Friedman, Yitao Liang, and Guy~Van den Broeck.
\newblock A semantic loss function for deep learning with symbolic knowledge.
\newblock In {\em International Conference on Machine Learning}, 2017.

\bibitem{Yao2023Tree}
Shunyu Yao, Dian Yu, Jeffrey Zhao, Izhak Shafran, Thomas~L. Griffiths, Yuan Cao, and Karthik Narasimhan.
\newblock Tree of thoughts: Deliberate problem solving with large language models.
\newblock {\em ArXiv}, abs/2305.10601, 2023.

\bibitem{Yao2022ReAct}
Shunyu Yao, Jeffrey Zhao, Dian Yu, Nan Du, Izhak Shafran, Karthik Narasimhan, and Yuan Cao.
\newblock React: Synergizing reasoning and acting in language models.
\newblock {\em ArXiv}, abs/2210.03629, 2022.

\bibitem{Zhang2023ImprovedLR}
Hanlin Zhang, Jiani Huang, Ziyang Li, M.~Naik, and Eric~P. Xing.
\newblock Improved logical reasoning of language models via differentiable symbolic programming.
\newblock {\em ArXiv}, abs/2305.03742, 2023.

\bibitem{Zhang2023TractableCF}
Honghua Zhang, Meihua Dang, Nanyun Peng, and Guy~Van den Broeck.
\newblock Tractable control for autoregressive language generation.
\newblock {\em ArXiv}, abs/2304.07438, 2023.

\bibitem{Zhu2023Ghost}
Xizhou Zhu, Yuntao Chen, Hao Tian, Chenxin Tao, Weijie Su, Chenyu Yang, Gao Huang, Bin Li, Lewei Lu, Xiaogang Wang, Y.~Qiao, Zhaoxiang Zhang, and Jifeng Dai.
\newblock Ghost in the minecraft: Generally capable agents for open-world environments via large language models with text-based knowledge and memory.
\newblock {\em ArXiv}, abs/2305.17144, 2023.

\end{thebibliography}

\end{document}